\documentclass{article} 
\PassOptionsToPackage{numbers, compress}{natbib}
\usepackage[preprint]{neurips_2025}

\usepackage{hyperref}
\usepackage{url}
\usepackage{graphicx}
\usepackage[utf8]{inputenc} 
\usepackage[T1]{fontenc}    
\usepackage{hyperref}       
\usepackage{url}            
\usepackage{booktabs}       
\usepackage{amsfonts}       
\usepackage{nicefrac}       
\usepackage{microtype}      
\usepackage{xcolor}         
\usepackage{amsmath}
\usepackage{algorithmic}
\usepackage{algorithm}
\usepackage{graphicx}
\usepackage{paralist}
\usepackage{subcaption}
\usepackage{multirow}
\usepackage{amssymb}
\usepackage{mathtools}
\usepackage{amsthm}
\usepackage{upgreek}
\usepackage{comment}
\usepackage{wrapfig}

\usepackage[capitalize,noabbrev]{cleveref}
\usepackage{autonum}

\title{Optimizers Qualitatively Alter Solutions\\And  We Should Leverage This}

\author{
Razvan Pascanu\textsuperscript{1,2} \And
Clare Lyle\textsuperscript{1}\And
Ionut-Vlad Modoranu\textsuperscript{3}\And
Naima Elosegui Borras\textsuperscript{4}\And
Dan Alistarh\textsuperscript{3,5}\And
Petar Velickovic\textsuperscript{1,6}\And
Sarath Chandar\textsuperscript{2,7} \And
Soham De\textsuperscript{1} \And
James Martens\textsuperscript{1} 
}

\begin{document}

\footnotetext[1]{\small Google DeepMind, UK}
\footnotetext[2]{\small Mila, Qu\'ebec AI Institute, Canada}
\footnotetext[3]{\small Institute of Science and Technology Austria (ISTA)}
\footnotetext[4]{\small Technische Universit\"at Berlin, Germany \& BIFOLD Berlin}
\footnotetext[5]{\small Red Hat AI}
\footnotetext[6]{\small Cambridge University, UK}
\footnotetext[7]{\small Politechnique Montreal, Canada}

\maketitle

\begin{abstract}

Due to the nonlinear nature of deep neural networks, one can not guarantee convergence to a unique 
global minimum of the loss when using optimizers that rely only on local information, such as gradient descent. 
Indeed, this was a primary source of skepticism regarding the feasibility of neural networks in the early days of the field. 
The past decades of progress in deep learning have revealed this skepticism to be misplaced, and a large body of empirical evidence shows that sufficiently large networks following standard training protocols exhibit well-behaved optimization dynamics that converge to performant solutions.
This success has biased the community to use convex optimization as a mental model for learning,
leading to a focus on training efficiency --- either in terms of required iteration, FLOPs or 
wall-clock time --- when improving learning algorithms. 
We argue that, while this perspective has proven extremely fruitful, another perspective 
specific to neural networks has received considerably less attention: namely, that the choice of optimizer (or learning algorithm) not only influences the \textit{rate} of convergence, but also the \textit{qualitative properties} of the learned solutions.
Restated, the choice of optimizer can and will encode inductive biases and change the \emph{effective expressivity of a given class of models}. 
Furthermore, we believe that the choice of the 
optimizer can be an effective way of encoding desiderata in the learning process. We contend that the community should
aim at understanding the biases of already existing methods,  
as well as aim to build new learning algorithms with the explicit intent of inducing certain 
properties of the solution, rather than solely judging them based on their convergence rates.
We hope that our arguments will inspire research to improve our understanding of how the learning process can impact the 
type of solution we converge to, and lead to a greater recognition of learning algorithm design as a critical lever that complements the roles of architecture and data in shaping model outcomes.

\end{abstract}

\section{Introduction}

Neural network research has a long and rich history, going back at least as far as \citet{mcculloch43a}. This journey has been marked by cycles of intense optimism followed by periods of skepticism regarding their efficacy, famously exemplified by the critique of the limitations of perceptrons to learn the XOR problem by~\citet{Min69}.
The current success of these models, starting with works like~\citep{hinton2006,bengio2006,ranzato2007} and the pivotal result on ImageNet by~\citet{alexnet2012},
rely heavily on the adoption of gradient based learning rules, made possible by the  back-propagation algorithm~\citep{rumelhart1986learning}.
Succinctly, modern machine learning relies on an iterative local learning rule, updating model parameters along a descent direction given by the gradient, optionally preconditioned by a matrix $P$.

Given the locality of the approach, and the non-convex nature of the neural network, a natural concern 
is: \emph{what kind of guarantees can one obtain in terms of the optimality of the solution}?
Interestingly, this was among 
the main themes in the critique around the XOR problem put forward in~\citep{Min69}; it resurfaced in 
different forms in the 90s and early 2000s, as a source of skepticism towards connectionist approaches to machine learning. For further discussion please see~\citep{Tesauro92, BengioLecun2007:scaling, Bengio09}. The field approached this question empirically, 
with initial successes coming from 
layer-wise pretraining~\citep{hinton2006, bengio2006}, which was assumed to initialize the model into a \emph{good basin of attraction}
as argued by the careful study of~\citep{erhan10a}. Shortly thereafter, works 
such as~\citep{Martens10, glorot10a} showed that good initialization and careful choice of optimizer can lead to 
strong performance when training these models, despite the persistent worry of \emph{bad local minima}. This has now become standard training protocol, achieving remarkable success.

\begin{figure}[t!]
    \centering
    \includegraphics[clip, trim=0cm 4cm 0cm 0cm,width=.7\textwidth]{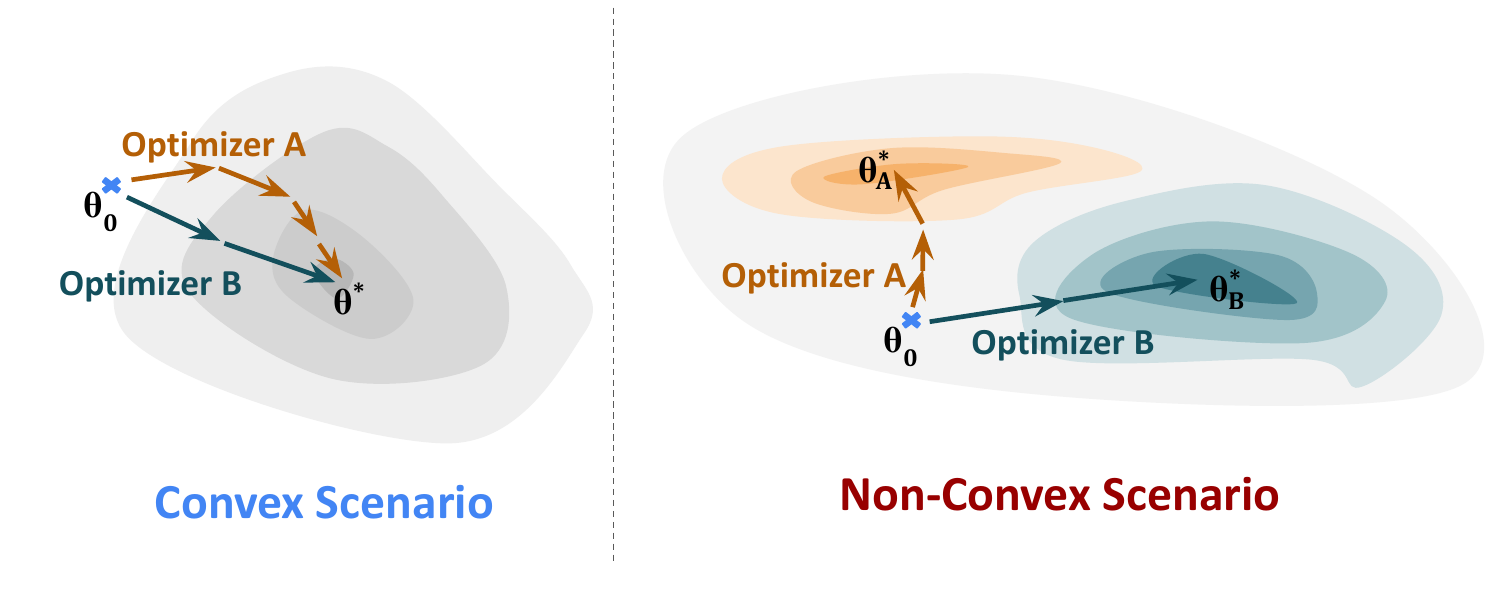} 
    \caption{Compared to the convex case, where optimizers can converge to a global minima, in the non-convex case (e.g.~for neural networks), different optimizers can lead training to converge to different minima. We argue that for neural networks, the choice of optimizer can lead to qualitatively different kinds of solutions, and that one can effectively leverage this as a mechanism for introducing inductive biases in learning, similar to architecture design.}
    \label{fig:high_level_idea}
\end{figure}

Follow-up work tried to formulate theoretical arguments for these observations, and study this behavior empirically, e.g.~\citep{Dauphin14,Choromanska15,GoodfellowV14,zhang2017understanding, li2018visualizing,Soltanolkotabi19}. These works suggested that, at scale, \emph{the loss landscape becomes well behaved, almost convex},
and that all local minima lead to similar performance to global minima. This included 
theoretical arguments borrowed from statistical physics~\cite[e.g.][]{Dauphin14} or 
empirical studies 
of how loss varies when interpolating between initialization and the convergence point, showing a \emph{monotonically decreasing} behavior~\citep{GoodfellowV14}, akin to the convex case. Note that the result of~\citet{GoodfellowV14} was later questioned by~\citet{frankle2020revisitingqualitativelycharacterizingneural}, however the original impact of the work on how learning was perceived cannot be ignored.  Local minima in fact often form linearly-connected basins~\citep{vlaar2022what}, where a convex combination of two local minima will also be an (approximate) local minimum. It was also shown in \citep{rame2023rewarded} that it is possible to linearly interpolate between parameters fine-tuned towards different task reward functions without degrading average performance on these tasks. However, recent works added nuance to this perspective, arguing that learning is actually split into at least two stages, a short initial stage in which learning jumps from one basin of attraction to another~\citep{keskar2017on}, and a second much longer stage wherein learning works within a locally convex region. Learning can fall within the \emph{lazy regime}, characterized by the Neural Tangent Kernel (NTK) \citep{jacot2018neural}, where model parameters remain close to initialization \cite{chizat2019lazy}, and, conversely, in the \emph{rich feature-learning regime}, characterized by models achieving generalization capabilities \citep{pmlr-v125-woodworth20a}.

While our understanding of learning dynamics has improved over time, such initial results biased the community towards not only  confidently borrowing ideas from 
the convex optimization literature and applying them to neural networks, but also on overly emphasizing the 
importance of convergence speed---the standard metric in convex optimization---as the main target when developing new learning algorithms. In some sense, the role of the 
optimizer is typically seen as being to exploit local convexity and
reach as fast as possible the minima of the basin of attraction the model finds itself in. The early literature 
providing the belief that generally for a sufficiently large model the basin of attraction the model finds itself in due to initialization and used protocols leads to a performant solution, and is in some sense sufficient, or \emph{good}.
However, the non-convex nature of neural networks means that the choice of optimizer can induce different paths in the parameter space and guide the transient stage of learning when the model jumps between basins of attraction. Different choices of optimizer can thus lead to different minima, which can have qualitatively different properties. 

\textbf{Our position is that the choice 
of optimizer itself provides an effective mechanism to introduce an explicit inductive bias in the process, and as a community we should attempt to understand it and \emph{exploit it} by developing optimizers aimed at converging to certain kinds of solutions. } 
The additional implication of this stance is that the optimizer can and does affect the effective expressivity 
of the model class (i.e. \emph{what solutions we can learn}). We argue that \emph{expressivity arguments that solely focus on the architecture design and/or data do not provide a complete picture} and could be \emph{misleading} for example if used to do model selection. The learning algorithm and choice of optimizer are also critical in shaping the characteristics of what are reachable functions and implicitly the final learned model.

Figure~\ref{fig:high_level_idea} provides a diagram of this position. On the left, convexity of the objective function causes the choice of optimizer to only affect speed of convergence. On the right, due to the different
paths in parameter space that different optimizers take, the process can converge to different minima.
The color and slight different shape of the minima is meant to suggest that these different solutions $\theta^*_A$ and $\theta^*_B$ do behave differently in some qualitatively meaningful way. 

\section{Looking at the inductive bias of learning}

The question of how the learning process or optimization impacts the learned solution has been explored 
previously, though maybe without sufficiently accentuating the role of the optimizer. 
In the seminal work of~\citet{Belkin_2019}, the authors try to understand the 
counter-intuitive observation that neural networks tend to behave better, and converge 
to better solutions, as they grow in size, a phenomenon  
referred to as \emph{double descent} (see also~\citet{Nakkiran19} for empirical evidence at scale). 
\citet{Belkin_2019} argues that size acts as a regularizer towards \emph{smooth} solutions, which---due to 
Occam's Razor---should generalize better. An alternative view is that scale of the system leads to 
\emph{small norm} solutions~\citep[e.g.][]{advani2017highdimensionaldynamicsgeneralizationerror,arora2019implicitregularizationdeepmatrix,neyshabur2018the}, provided that the architecture is initialized to small values. Intuitively, one can internalize this effect by acknowledging that increasing the model size will lead to an exponential growth in the number of critical points of the loss and specifically of the number of minima. Therefore the likelihood that gradient descent will converge to a minima \emph{nearby} the initialization grows considerably, leading to finding solutions of small norm given that initialization is close to $0$.  This is becoming 
a popular view in learning theory~\citep{Zhu19}, whereas the optimization process limits the class of functions considered, therefore allowing for  better generalization bounds. 

While the focus in these works is on the role of \emph{scale}, we want to emphasize that the phenomenon is predicated on the use of \emph{gradient descent} as the optimizer. That is, the use of a different learning algorithm -- which for example can take very large steps in the parameter space or not be a form of local search --  will not be biased towards converging to a nearby minima, therefore eliminating the regularization effect of scale. 

Another thoroughly discussed topic in the literature is that of \emph{flat} vs. \emph{sharp} or \emph{narrow} minima~\citep{Hochreiter97Flat}. Empirically it has been observed that models trained with  \emph{stochastic} gradient descent generalize better compared to those trained by \emph{batch} gradient descent~\citep[e.g.][]{bottou03, bottou07, keskar2017on}. The theoretical argument for this discrepancy, as presented in~\citet{Hochreiter97Flat}, is that the noise of the stochastic gradient method prevents convergence to a \emph{narrow} minima, and, relying on a minimum description length argument, this suggests that  \emph{flatter} minima will generalize better. This argument is in line with our proposal, and has led to development of  optimization algorithms like SAM~\citep{foret2021sharpnessaware} that are meant to \emph{improve generalization rather than convergence speed}. Additionally,~\citet{fort2022drawingmultipleaugmentationsamples} showcase that the regularization effect coming from the optimizer can not easily be replaced by an explicit gradient-agnostic term. This argument is in line with the success of SAM over additive regularization terms aimed at imposing convergence to flat minima~\citep{Chaudhari16} or methods that add noise in the optimization process to improve exploration~\cite{Neelakantan2015AddingGN}. Also worth  noting is the discussion around whether the concept of \emph{sharp/narrow} or \emph{flat} minima is sufficiently well defined~\citep{dinh2017sharpminimageneralizedeep}, particularly in the case of scale-invariant networks~\citep{kwon2021asam}.

As we will argue in the rest of the paper, we believe that lines of research such as these have not received enough attention from the community and have overly focused on \emph{in-domain generalization} without expanding to other properties of the solution. 

Building on the flat/narrow minima discussion, a slightly different perspective~\citep[e.g.][]{barrett2022implicitgradientregularization,smith2021originimplicitregularizationstochastic} is that 
gradient descent  has an \emph{implicit form of regularization} that goes beyond the noise introduced by 
sampling the data, 
and this regularization is present in both \emph{batch} and \emph{stochastic} variants of gradient descent, 
and differs between the two. 
Furthermore, \citet{bernstein2024oldoptimizernewnorm} exemplifies how the different optimizer can be understood as gradient descent under a different choice of norm, which implies that the implicit regularization effect of the algorithm operates under different choices of norms as well for these different optimizers, therefore representing different inductive biases.

\citet{amari2020preconditioning} explores the role of the preconditioner in the ability of an architecture 
to generalize. Specifically, the work brings into question the assumption that second order methods hurt 
generalization, and argues that in the case of noisy labels, the use of a preconditioner should lead to better 
performance, while a first order optimizer will perform better in the noiseless scenario. 

\paragraph{Our thesis.} Our argument here is two-fold. First, we want to emphasize more widely the perspective that the optimizer or learning algorithm, 
similar to all other components of the deep learning pipeline, is a rich source of inductive bias in learning.  In other words, we argue to expand from (in-domain) generalization, and argue that the optimizer can be an \emph{effective and generic vehicle for various inductive biases}, that can relate to other properties of the solution, such as sparsity, structure of the representation, robustness to catastrophic forgetting, and so on. By modulating the updates of an iterative learning algorithm, we are altering the credit assignment mechanism by which learning decides which weights get blamed for what part of the loss. This is what shapes up the representations of the system, 
and defines the type of solution learned. 

Second, our objective 
is to emphasize that the optimizer or learning algorithm has a non-trivial impact on the expressivity of the selected architecture, a fact typically ignored in the literature, 
where the expressivity of the class of functions considered tends to be a deciding factor 
in model selection. We believe there is a duality between architecture design and 
optimizer design, and while the community has been heavily biased towards
architecture design, we want to argue that at least it is worth considering if certain desiderata might not be easier to obtain via altering the 
learning algorithm. This perspective becomes considerably more 
important when dealing with large pretrained models, as it is becoming more of a norm, \emph{where changing the architecture of the pretrained model to encode some bias is not an option, however changing the learning algorithm used to finetune the system is}.

\section{Examples of qualitative different minima due to the optimizer}

Before expanding our arguments further, we will present a few test cases that we believe exemplify the potential of the optimizer to impact the learned solution beyond improving its ability to generalize. The aim for 
these examples is not to act as a methodological contribution, but rather they are more akin to thought experiments that will 
help us formulating our argument. The hope is that by being more concrete in the argumentation, the community can respond more directly to our position.

\subsection{Non-diagonal preconditioners, Catastrophic Forgetting and Forward Transfer}

One particular topic of interest in the recent literature is learning under non-stationary settings, presented in 
different formulations within the Continual Learning Problem (e.g. task incremental, class incremental, task-agnostic, etc)~\citep[e.g.][]{Bing17, Parisi18review, Lange22, hadsell2020embracing}.
Among the main phenomena of study in these settings range from 
\emph{catastrophic forgetting}~\citep[e.g.][]{robins1995catastrophic, french1999catastrophic, kirkpatrick2017overcoming}, to \emph{forward transfer}~\citep{hadsell2020embracing} which can be understood either as \emph{plasticity}~\citep{lyle2021understanding, nikishin2022primacy, dohare2021continual} or fast adaptation~\citep{finn2017model}.
Several architecture modifications and regularization terms, alongside replay based methods have been proposed to address these issues.

\begin{figure}[t!]
    \centering
    \includegraphics[width=.7\textwidth]{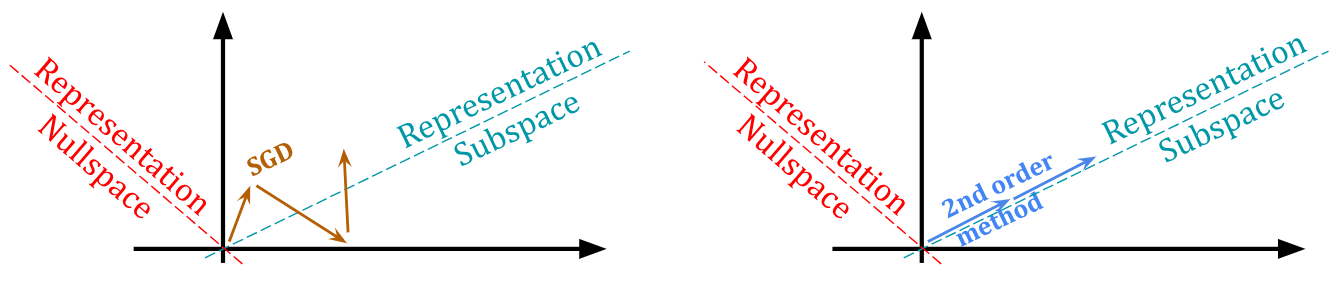} 
    \caption{Diagram depicting the intuition of why second order method lead to more localized representations. Note how updates of SGD move within the entire space, inadvertently leading to a representation that occupies a larger space, while a second order method avoids wasteful movement staying within a smaller subspace. }
    \label{fig:wasteful}
\end{figure}

In this subsection however we will take a different view. We will try to reason through what is the impact of the optimizer. Particularly we will look at the impact of non-diagonal preconditioners within these continual learning problems. There is a rich literature on using second order methods 
for improving convergence speed, from algorithms like Natural Gradient~\citep{Amari98}, Hessian-Free~\citep{Martens10}, K-FAC~\citep{martens15}, Shampoo~\citep{gupta2018shampoopreconditionedstochastictensor} or WoodFisher~\citep{singh2020woodfisherefficientsecondorderapproximation}. At the core of these methods, an approximation of the Hessian or Fisher Information Matrix is used to precondition (i.e. multiply) the gradient in order to correct by how \emph{quickly} it changes as parameters move~\citep{Nocedal06} --- i.e. if the gradient is not changing when you move in a certain direction then you have low curvature and you can afford to take a large step, otherwise you have to take a small one.

In turn, the gradient tries to estimate \emph{independently} for each parameter, what is the impact on the loss for a small change $\Delta$ in the parameter value. In other words, the gradient on weight $\theta_{i}$ is saying --- under a local linearization of 
the objective --- whether the loss will increase or decrease if the parameter is being changed. Such a mechanism for doing credit assignment has deep implications in both stationary and non-stationary settings, and it has been argued as being one of the main culprits behind issues like catastrophic forgetting~\citep{hadsell2020embracing} by leading to \emph{tug-of-war} dynamics in learning.
But more importantly for our discussion, due to treating each parameter independently, learning typically 
\emph{over-shoots}, and if there are multiple directions in parameter space along which the loss can be reduced, gradient descent will move along all. This over-shooting and follow-up correction can be seen as \emph{wasteful movement}, that a second 
order method would aim to avoid. 

In this context, see Figure~\ref{fig:wasteful}, we can interpret these wasteful movement as 
perturbations in the representations learned by the system. And if we assume that these 
representations live in a subspace (or on some manifold), these perturbation will move 
them off the manifold, increasing the dimensionality of the subspace they will end up occupying. We believe that over time such perturbations lead to the model learning to waste capacity, using a larger subspace then necessary to encode information.

However, a non-diagonal form of a second order preconditioner, within its \emph{off-diagonal} elements, captures 
how much a change of weight $\theta_i$ will affect the gradient of $\theta_j$. That means that the learning now 
can take into account, to a certain degree, the correlated effect of moving along different axes.  When correcting by the preconditioner (i.e. correcting by the curvature), in an 
idealized setting, this \emph{wasteful movement} is eliminated.
We argue that the effect 
of this is to learn more \emph{localized} and less redundant representations, that occupy 
a lower dimensional subspace that those learned by gradient descent. We believe 
that our argument applies specifically to \emph{non-diagonal} preconditioner, which are 
the only elements that capture the relationship between different entries of the gradient. And while 
\emph{distributed} and \emph{redundant} representations can be desirable, it has been 
previously argued~\citep[see][]{aljundi2019selflesssequentiallearning, schwarz2021powerpropagationsparsityinducingweight} that learning more localized and compressed representation for tasks is very beneficial for continual learning.
The intuition being that by using less of the capacity of the model, future learning will induce less interference, and hence less 
forgetting. The work of \citep{doan2021theoretical} analyzes forgetting from a theoretical viewpoint by the NTK overlap matrix, showing that higher accuracy and lower forgetting is given by lower task overlap.

Additionally, when learning the subsequent task with a second order 
optimizer, the learning process has \emph{more degrees of freedom}, and 
can move with equal efficacy in more directions than gradient descent, 
because of the correction of the gradient. Additionally, the lack of 
over-shooting also makes it less likely for the update to destroy previously learned information. 
Overall, having an effective larger 
degree of movement and more precise step sizes means that the likelihood of interference will be less.

\begin{wrapfigure}{l}{.7\textwidth}
    \centering
    \includegraphics[width=.5\linewidth]{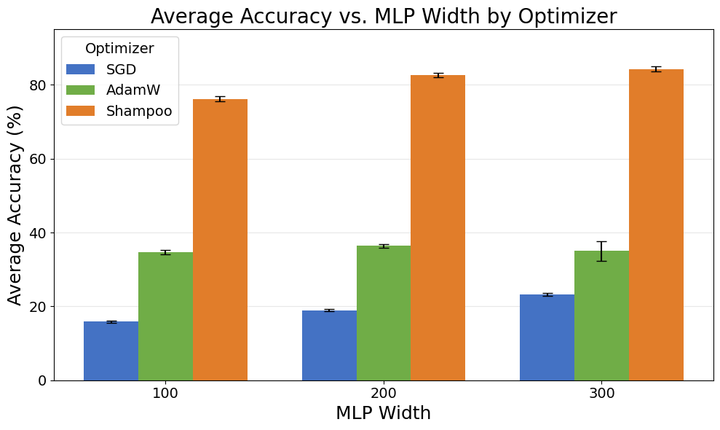}~\includegraphics[width=.35\linewidth]{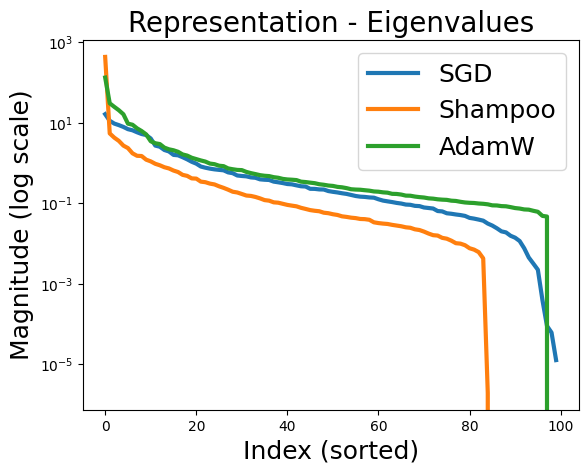} 
    \caption{\textbf{Left:} Test accuracy averaged over 3 variants of permuted MNIST learned sequentially with different optimizers (for different sizes of MLPs); \textbf{Right:} Spectrum of the covariance of the representation for the 100 units MLP. Note how the number of significant eigenvalues (and effective rank) of the Shampoo trained model is lower.}
    \label{fig:results}
\end{wrapfigure}
To illustrate this behavior we look at the representation of a small single hidden layer MLP when learning three different permutations of MNIST sequentially. We use SGD, AdamW~\citep{adamw}, which employs a \emph{diagonal preconditioning}, and Shampoo~\citep{gupta2018shampoopreconditionedstochastictensor} which can be thought of as a computationally efficient non-diagonal counterpart of Adam.  Given the difference in convergence speed, and to avoid artifacts from overfitting, we tune hyper-parameters independently for each algorithm, using the same batch size and early stopping based on test error to decide when to stop learning each task. In Figure~\ref{fig:results} (left) we show average test error\footnote{averaged over the three tasks, and 10 different random seeds; note that for each trial we use exactly the same initialization for the three algorithms} on all tasks at the end of training, highlighting that Shampoo is outperforming the others. Note that due to early stopping based on test performance on each task, this advantage comes from the system \emph{not forgetting previously learned tasks}. Figure~\ref{fig:results} (right) we show the spectrum of the covariance of the representation, highlighting that the model trained with Shampoo, even when initialized at exactly the same point as that trained with SGD or AdamW, has a lower effective rank (i.e. their representation occupies a lower-dimensional subspace) as outlined by our intuition. This suggest that indeed by changing optimizer, there is a qualitative impact on representations, which becomes more localized or of lower effective rank. 

\begin{figure}
    \includegraphics[height=3.25cm]{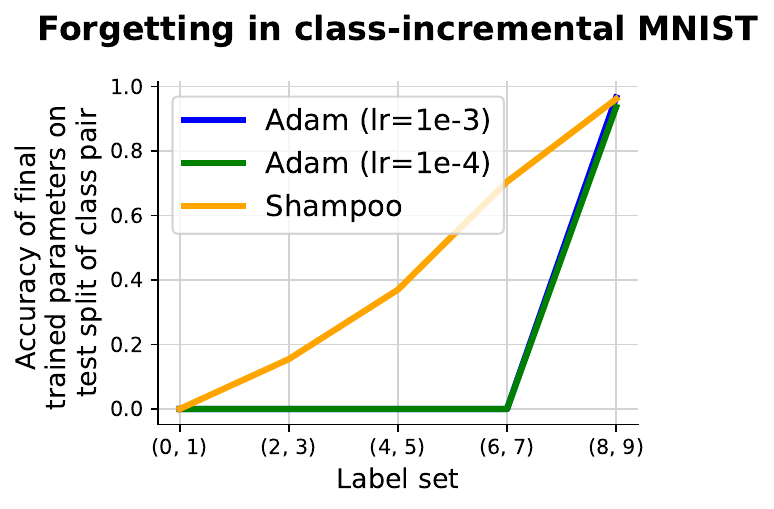}
    \includegraphics[height=3.25cm]{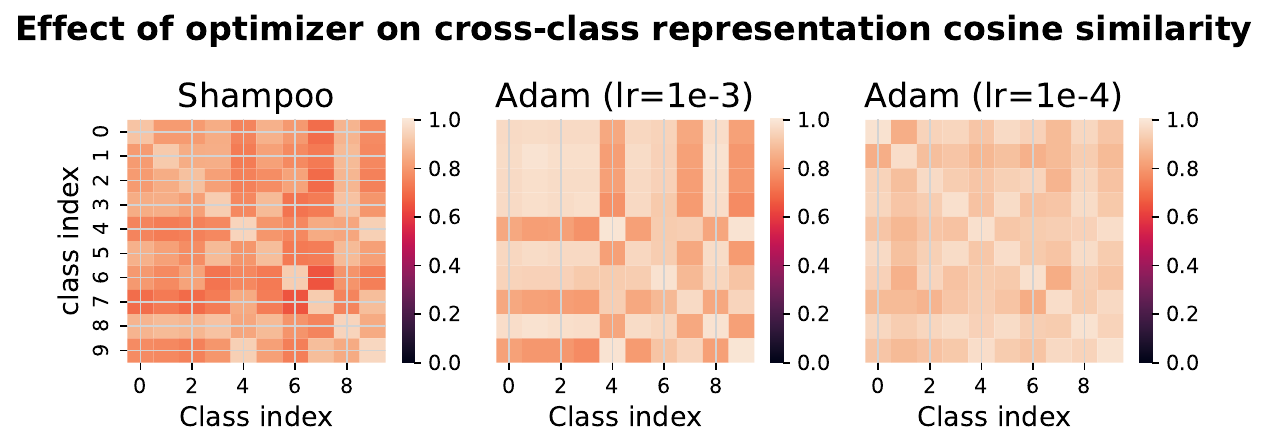}
    \caption{\textbf{Left:} catastrophic forgetting in a 2-layer MLP trained on class-incremental MNIST, where the network trains on each pair of classes sequentially. All networks exhibit worse performance on earlier class pairs, but the decline in performance is much sharper for Adam than for Shampoo. This effect is not mitigated by reducing the learning rate on Adam. Right: visualization of the alignment between features of different classes in each network. Features are more degenerate (higher cross-class cosine similarity) when training with Adam than with Shampoo.}
    \label{fig:shampoo-incremental}
\end{figure}
To further illustrate this point, we consider the effect of the Shampoo optimizer on a related problem where pairs of MNIST digit classes are shown sequentially to the learner: for example, the network is first trained for one epoch on images of the digits 0 and 1, then this data is discarded and the network continues training for one epoch on images of the digits 2 and 3, and so on. We train a network with either the shampoo optimizer or Adam and then take the final parameters at the end of training for evaluation. We see in Figure~\ref{fig:shampoo-incremental} that the network trained with Adam is only able to attain a high accuracy on the final pair of classes on which it was trained. Shampoo, while still exhibiting some forgetting, attains nontrivial accuracy on previous class pairs, demonstrating reduced interference between classes.
Looking more closely, we observe that Shampoo's resilience to forgetting is accompanied by reduced interference between representations of inputs corresponding to different classes, suggesting that the pre-conditioning performed in the Shampoo \textit{update} translates to improved conditioning of the learned \textit{features}.

Lastly, it is worth also mentioning the potential impact of optimizer on other aspects of continual learning. \citet{lyle2021understanding}
show that \emph{loss of plasticity} can be understood from an optimization perspective, being caused by ill-conditioning of the learning process. In particular, works which have aimed to mitigate loss of plasticity have done so by modifying the optimization process to preserve various properties of a randomly initialized network thought to facilitate learning~\citep{lewandowski2023directions, kumar2024maintaining, lyle2024normalizationeffectivelearningrates}. While certain network pathologies such as dormant ReLU units~\citep{sokar2023dormant} introduce challenges that a second-order method can not trivially fix, 
they are not the only mechanism to lose plasticity~\citep{lyle2024normalizationeffectivelearningrates} and their effect is still ameliorated by using stronger optimizers. Therefore 
the \emph{stability-plasticity} trade-off \textbf{depends} on the choice of optimizer. This is contrary to theoretical treatments of this question in the field, that typically focus purely on expressivity arguments, ignoring the learning process~\cite{kumar2023continuallearningcomputationallyconstrained}.

\subsection{Preconditioners, plateaus and sparsity}

In our next example we will exploit the "duality"\footnote{Note that we use the term duality in an informal manner. While we hypothesize that many aspects of architecture design can be cast in terms of changes to the learning rule, establishing whether there is a duality or not requires considerable more work and thought} between reparametrization and optimization to recast an existing work, 
the Power-propagation algorithm~\citep{schwarz2021powerpropagationsparsityinducingweight}, 
as a particular choice of preconditioner that forces learning to 
favor \emph{sparse solutions}. We argue that the optimization perspective of this algorithm is actually more natural, and easier to 
apply in practice. Our main goal is to highlight that one can build 
optimizers that \emph{sacrifice convergence speed in order to bias learning towards certain kinds of solutions}.

Power-propagation, proposes a reparameterization of neural network
where $\theta$ gets replaced by $\phi|\phi|^{\alpha -1}$, for some $\alpha > 1$. I.e., ignoring sign issues for simplicity, we raise 
each parameter to some power $\alpha>1$ before using the weights in the computational graph representing the model. 

The intuition, described at length in the original work, is that 
 the reparameterization introduces, along each dimension corresponding to 
different parameters $\theta_i$, a saddle centered at $0$ in the original 
loss surface.  This can easily be seen by noting 
that the gradient w.r.t.~$\phi$ will be multiplied by $\phi$. To simplify notation, let us ignore the 
absolute value formulation, and consider 
a simpler form, where $\theta$ is replaced by $\phi^\alpha$, ignoring the impact it might have on the sign of the weight. In this simplified scenario the gradient becomes:
$
 \frac{\partial \mathcal{L}}{\partial \phi} = \alpha\frac{\partial \mathcal{L}}{\partial \phi^\alpha} \phi^{\alpha-1}.
$

\begin{wrapfigure}{l}{0.25\textwidth}
    \centering
    \includegraphics[width=.98\linewidth,trim={1.2cm 2.5cm 0 2cm},clip]{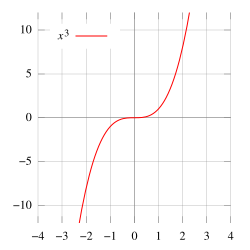} 
    \caption{Saddle point created by using $\phi^\alpha$, for $\alpha=3$. Note that we chose an odd power to use the simplified formula $\phi^\alpha$ which is more intuitive to reason about.}
    \label{fig:saddle}
\end{wrapfigure}

The shape of the loss with respect to a specific parameter $\phi_i$ around $0$ will roughly look like the depiction in Figure~\ref{fig:saddle}. The work then proceeds to argue that these saddles will make the learning process 
less likely to converge to solutions for which many parameters move away from $0$. The reason being that for a parameter $\theta_i$ to escape $0$, it requires many gradient steps. So learning 
will more likely make use of other parameters that are easier to change to reduce the loss. This effect can additionally be amplified by adding $L_2$ norm regularization on the weights, typically present in most learning protocols. 
Given that the learned solution has few parameters with large magnitude, 
and many which are close to $0$, the model can be effectively sparsified by simply thresholding, with minimal impact on performance. 
The original work shows 
ample evidence of this being true, both by looking at the distribution of weight magnitudes after training, highlighting that it becomes sharper, but also by effectively sparsifying trained models on typical benchmarks with and without the proposed reparameterization.

We argue that the parametrization itself has the goal of changing the dynamics of the optimizer, and 
an equivalent effect can be obtained by directly changing the preconditioner, without reparametrizing the model. 
For this particular case, reframing the algorithm as a change of 
preconditioner is also more advantageous. During training it does not 
require exponentiation of the parameters for the forward pass, saving FLOPs, but more importantly the original method required the optimizer 
not to properly correct the gradient by curvature. As outlined in~\citep{schwarz2021powerpropagationsparsityinducingweight}, if 
Power-propagation is blindly used with the Adam optimizer, the efficacy of 
the method drops considerably. The work proposes an alternative optimizer
which corrects by the curvature of the original loss, but not by the curvature introduced by the reparametrization. For the method to converge to \emph{sparse} solutions, learning needs to get stuck in these saddles and allow 
some parameters to be more flexible than others, depending on their magnitude, forcing the optimizer to operate on a lower dimensional subspace. 

Given that the reparameterization also requires changing the optimizer, we propose here that equivalent dynamics could be obtained by \emph{not reparametrizing the model and only changing the optimizer}.  If the 
original model would be optimized using a preconditioned SGD, with 
preconditioner matrix
$P = diag (|\theta|^\beta)$
with $\beta>0$, this leads to identical learning dynamics. We can see this by noting that the saddles in the original method were
introduced by having the gradients be scaled by $\phi^{\alpha-1}$ and  
our new proposed conditioner has an equivalent step-wise behavior\footnote{When taking the derivative in the reparameterization, we get the gradients are also scaled by $\alpha$ -- we fold this into the learning rate}.  

Intuitively, updates will become vanishing small for small magnitude weights, and learning will be unable to move them unless there is no other way to minimize the overall objective. Note that the effect is 
different from a traditional \emph{additive} regularizer. Therefore 
we would argue that the exact effect of this method can not be replicated by a standard \emph{additive} regularization scheme.

This suggests, as argued by our position, that one can choose an optimization algorithm, or rather choose a preconditioner 
to \emph{enforce sparsity, a qualitative property of the solution. And that this choice can be more attractive even if it implies slowing down convergence}.
This is discussed in the Power-propagation paper, were optimization is less well behaved and one needs to carefully tune $\beta$ (or $\alpha$ in the original work) and learning rate in order to still ensure convergence to solutions 
that are equally performant. However the impact on the ability to sparsify the model is considerable, leading to state of 
art results.

\section{Our perspective and its limitations}

In~\citet{griffiths2020understandinghumanintelligencehuman}, the author 
argues that the characteristics of human intelligence are defined 
by \emph{biological limitations} that artificial systems do not have. To make this intuition clearer we can consider the well known game between AlphaGo~\cite{silver2016mastering} against Lee Sedol, and in particular the unexpected \emph{move 37}, which was at the time referred to as alien-like. One can argue that the reason that this move seemed that way is because AlphaGo minimizes directly the 
overall objective, namely to win the game. In contrast, due to the large search space and limited ability to carry parallel computations and store large amounts of information, humans need to find a suitable decompositions of the problem into sub-goals that are more accessible and then to re-compose these partial solution into a complete one. Since move 37 did not try to solve any potential sub-goal, it seemed alien. 

Taking a step back, this line of thought suggests that compositionality, 
a core aspect of human intelligence, is not always optimal --- or at least that there are solutions that do not involve compositionality. In fact it 
often might lead to suboptimal solutions. However it provides alternative advantages, as for example  fast adaptivity or making \emph{infinite use of finite means}. All of these properties are important for a system that interacts with an ever-changing environment, where the ability to generalize out-of-distribution, assimilate new knowledge fast and be able to transfer from one setting to another is vital. 

This suggest that an end-to-end learning scheme trying to minimize a fixed objective has little incentive to discover such compositional structures or representations, partially because they might not be optimal. Indeed they may incur a price in performance and are, for humans, a by-product of inductive 
biases given, according to~\citet{griffiths2020understandinghumanintelligencehuman}, as limitations to the learning process. 

Furthermore, the concept of generalizing \emph{out-of-distribution}, under different choices of what out-of-distribution means, is crucial 
to obtain general intelligent systems, and compositionality provides only one mechanism to generalize this way. Algorithmic reasoning (or the ability to imitate or execute algorithms) can be another mechanism that allows certain forms of OOD generalizations, e.g. when it comes to concepts like causal reasoning or mathematics. It has been argued that learning algorithms requires new inductive biases in the learning process, typically referred to as \emph{algorithmic alignment}~\cite{nerem2025graphneuralnetworksextrapolate}. Let us consider the task of adding numbers. We would like our systems to discover the \emph{algorithm of adding two numbers}, in order to generalize to any numbers, not merely to find shortcuts or representations that it allows it to operate in some finite range. However algorithms, and in particular traces of algorithms have very different characteristics from the internal mechanics of a neural networks: they rely on localized representation, sparse access and sparse edits into the representation~\citep{dudzik2024asynchronousalgorithmicalignmentcocycles}. These differences are part of the cause of why learning the underlying algorithms is difficult, as the current architectures do not trivially have these biases.

We argue therefore that providing inductive biases into the learning process is still crucial and one of the more important problem of existing systems. The \emph{end-to-end learning} mantra will not be able, on its own, to discover solutions that have the necessary properties for many settings of interest, like compositional structure, able to exactly represent and discover algorithms, generalize to new distributions and so forth. Throwing more data will not fundamentally solve the problem either.

Inductive biases have been typically explored through architectural changes or by curating data that the model trains on. While this had led to successful mechanisms to induce certain properties, like rotation invariance or translation invariance in vision systems, our main argument is that the learning rule in general, and the choice of preconditioner in particular can have an equal impact. And that development of new optimizers has been overly focused on convergence speed which biased the field towards diagonal methods or certain type of approaches that can scale and find a good balance between computational cost and speed ups in convergence. While this on its own is not a bad thing, and the community should continue to explore optimizers from this perspective,  if we 
focus on developing optimizers as a vehicle for various inductive biases, this can lead to new insightful results. 

\begin{figure}[t!]
    \centering
    \includegraphics[width=.7\linewidth]{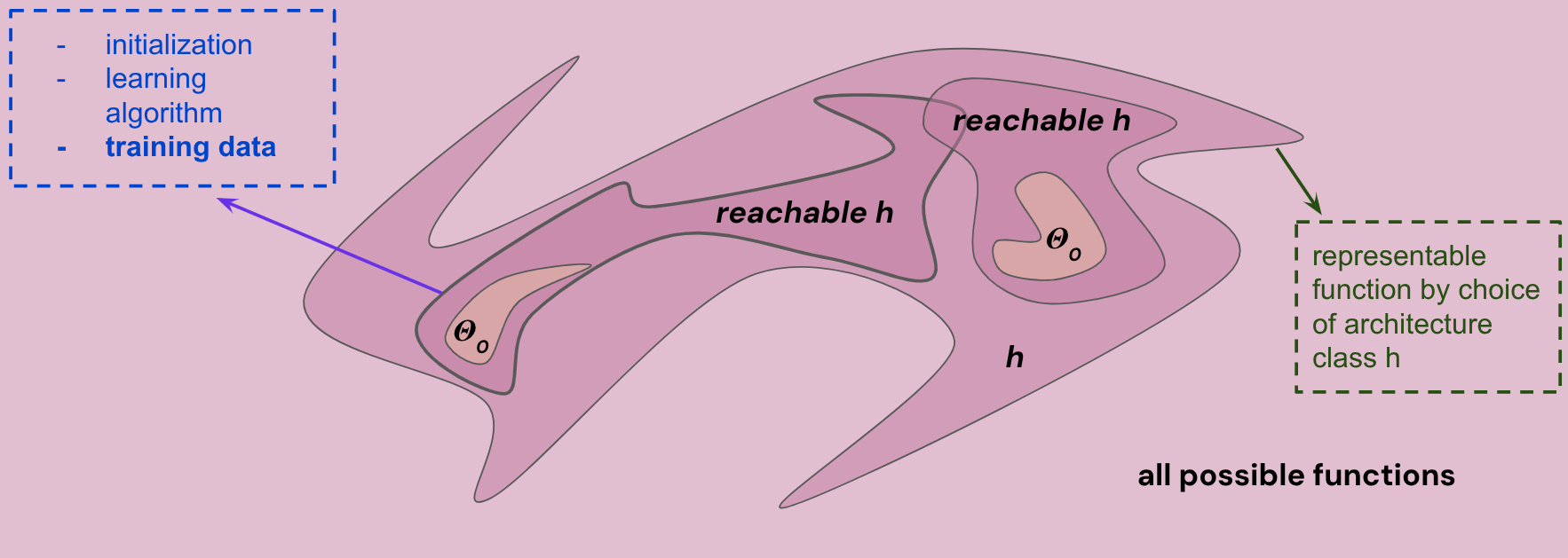} 
    \caption{Diagram depicting our main position. While the choice of architecture and model size limits the set of functions that are realizable, the optimizer, as well as data and other aspects of the training protocol further limits the functions that are reachable through learning. This can be seen as learning process and optimizer being crucial to define the effective expressivity class of a system, but also that the optimizer can play a crucial and effective role in introducing inductive bias in learning by how it restricts and traverses the the set of reachable functions. }
    \label{fig:learning}
\end{figure}

Additionally, the main mechanism through which an optimizer provides an inductive biases is by further restricting the search space of possible functions that the class of model can represent. We refer to Figure~\ref{fig:learning} for a visualization of the argument. This points to the importance of considering, among other things, optimization in any expressivity argument, particularly when thinking about model selection. To give a specific example, we consider the question of \emph{Turing Completness}. In the literature, several works are actively debating whether Transformers are \emph{Turing Complete} or not~\citep[e.g.][]{perez21,veličković2024softmaxforsharpoutofdistribution}, implicitly making a comparison with RNNs, known to be Turing Complete since the 90s~\citep{hava92,hava21}. The argument usually is that for transformers to lead to AGI-like behavior, Turing completeness seems like a pre-requisite. However these are purely expressivity arguments that ignore whether learning can discover the target behavior. One could argue that due to the well known vanishing/exploding gradient problem~\citep{pascanu13} there are functions that an RNN can technically express, but that are not reachable from initialization using gradient based method as it requires traversing regions in which the gradient signal vanishes in non-trivial ways, leaving no learning signal for the model. This would imply that the reachable RNNs by gradient descent might in fact not be Turing Complete and the expressivity of this class might be quite different. 

\paragraph{Limitations of our position.} \emph{Counterargument 1.}The relationship between reparameterization and optimization, suggests that what is achievable 
by a choice of optimizer, can be equally well achieved by 
a reparameterization of the model. Which begs the question of why 
putting the inductive bias into the optimizer rather than directly into the model via reparameterization? Why is it not ok to fix one of the two choices in order to reduce the search space? Our view is not to dispute this statement, but rather to encourage exploration of both. The reasoning being that the change in perspective might make encoding certain inductive biases much easier or efficient. Additionally, there could be instances, like in the scenario of large pretrianed models, where one can not choose the architecture in order to encode additional inductive bias, but can choose the learning process used to finetune the system. 

\emph{Counterargument 2.} Our position also relies on the assumption that obtaining some desirable behaviors from a learned system can only be achieved via providing some inductive biases to the system. However hand-engineering these biases, or even enumerating them seems wrong!  To what extent do we believe that 
sparse and localized representation are really needed for certain forms of OOD generalization ? Is it not better to discover the solution purely from data and interactions with the world without cooking in any inductive bias ? Unfortunately, answering the question of what should be prescribed and what should be learned is far from trivial. Our view is that so far the systems that we use do have certain biases imposed by the choices we have made as a community, whether we are aware of them or not. Not relying on inductive biases and learning everything from the data is an unrealistic expectation. In this paper we are just trying to argue that since providing inductive biases is unavoidable, we might as well make them explicit, understand them and exploit them.

\section{Conclusion}

In this work we presented our position that optimizers and preconditioners should be also studied or explored  in order to encode inductive biases. We highlighted our point of view through two different examples. In the first example we argued that second-order optimizers using non-diagonal preconditioners lead to less interference and hence less forgetting. In the second example, we 
take inspiration from a published method, Power-propagation~\cite{schwarz2021powerpropagationsparsityinducingweight}, showing that it can be interpreted as a change in preconditioner rather than a reparametrization of the architecture. This 
particular example highlights how trading off training efficiency can lead to an optimizer that, beside minimizing the loss, can also lead 
to certain type of solutions, in this case more sparse ones. 

We believe that this provides a basis to argue that as a community 
we should explore or study optimizers also as a mechanism to encode inductive bias. We believe this perspective to be novel 
and to have the potential of leading to new interesting research directions from researchers working on optimization.
Furthermore, we argue that the choice of optimizer should be more on 
an equal footing with the choice of architecture, and that the 
interplay between the two should be further explored, as is for 
example the impact of the learning algorithm on the expressivity 
of a certain class of models.

\bibliographystyle{plainnat}
\bibliography{references}

\end{document}